\documentclass[runningheads]{llncs}
\usepackage{graphicx}
\usepackage{algorithm2e}
\usepackage{amsmath,amsfonts,amssymb}
\usepackage{bm}
\usepackage{booktabs}
\usepackage{graphicx}
\usepackage{hyperref}
\usepackage{url}
\usepackage{semantic}
\usepackage{pifont,stmaryrd}
\usepackage{subcaption}
\usepackage{todonotes}
\setuptodonotes{inline}
\usepackage{listings}

\usepackage{amsmath,amsfonts,bm}

\def\eqref#1{equation~\ref{#1}}

\def\1{\bm{1}}

\def\rvw{{\mathbf{w}}}
\def\rvx{{\mathbf{x}}}

\def\rvz{{\mathbf{z}}}

\def\vmu{{\bm{\mu}}}
\def\vphi{{\bm{\phi}}}

\def\vtau{{\bm{\tau}}}
\def\vtheta{{\bm{\theta}}}

\def\mI{{\bm{I}}}

\DeclareMathAlphabet{\mathsfit}{\encodingdefault}{\sfdefault}{m}{sl}
\SetMathAlphabet{\mathsfit}{bold}{\encodingdefault}{\sfdefault}{bx}{n}

\newcommand{\Ls}{\mathcal{L}}
\newcommand{\R}{\mathbb{R}}

\newcommand{\dom}[1]{\mathrm{dom}(#1)}
\newcommand{\cod}[1]{\mathrm{cod}(#1)}

\newcommand{\ty}[1]{\bm{\tau}_{#1}}

\newcommand{\mcC}{\mathcal{C}}

\newcommand{\mcO}{\mathcal{O}}
\newcommand{\mcS}{\mathcal{S}}

\newcommand{\cmark}{\ding{51}}
\newcommand{\xmark}{\ding{55}}

\begin{document}
\title{Computing with Categories in Machine Learning\thanks{Supported by the JST Moonshot Programme on AI Robotics (JPMJMS2033-02).}}
%
%
\author{Eli Sennesh$^1$, Tom Xu$^2$, and Yoshihiro Maruyama$^2$}
\authorrunning{E. Sennesh et al.}
%
\institute{Northeastern University, Boston, USA\\
\email{sennesh.e@northeastern.edu} \and
Australian National University, Canberra, Australia\\
\email{\{tom.xu,yoshihiro.maruyama\}@anu.edu.au}}
\maketitle              
\begin{abstract}
Category theory has been successfully applied in various domains of science, shedding light on universal principles unifying diverse phenomena and thereby enabling knowledge transfer between them. Applications to machine learning have been pursued recently, and yet there is still a gap between abstract mathematical foundations and concrete applications to machine learning tasks. In this paper we introduce DisCoPyro as a categorical structure learning framework, which combines categorical structures (such as symmetric monoidal categories and operads) with amortized variational inference, and can be applied, e.g., in program learning for variational autoencoders. We provide both mathematical foundations and concrete applications together with comparison of experimental performance with other models (e.g., neuro-symbolic models). We speculate that DisCoPyro could ultimately contribute to the development of artificial general intelligence.

\end{abstract}

\section{Introduction}
\vspace{-1em}

\begin{figure*}[b!]
    \includegraphics[width=\textwidth]{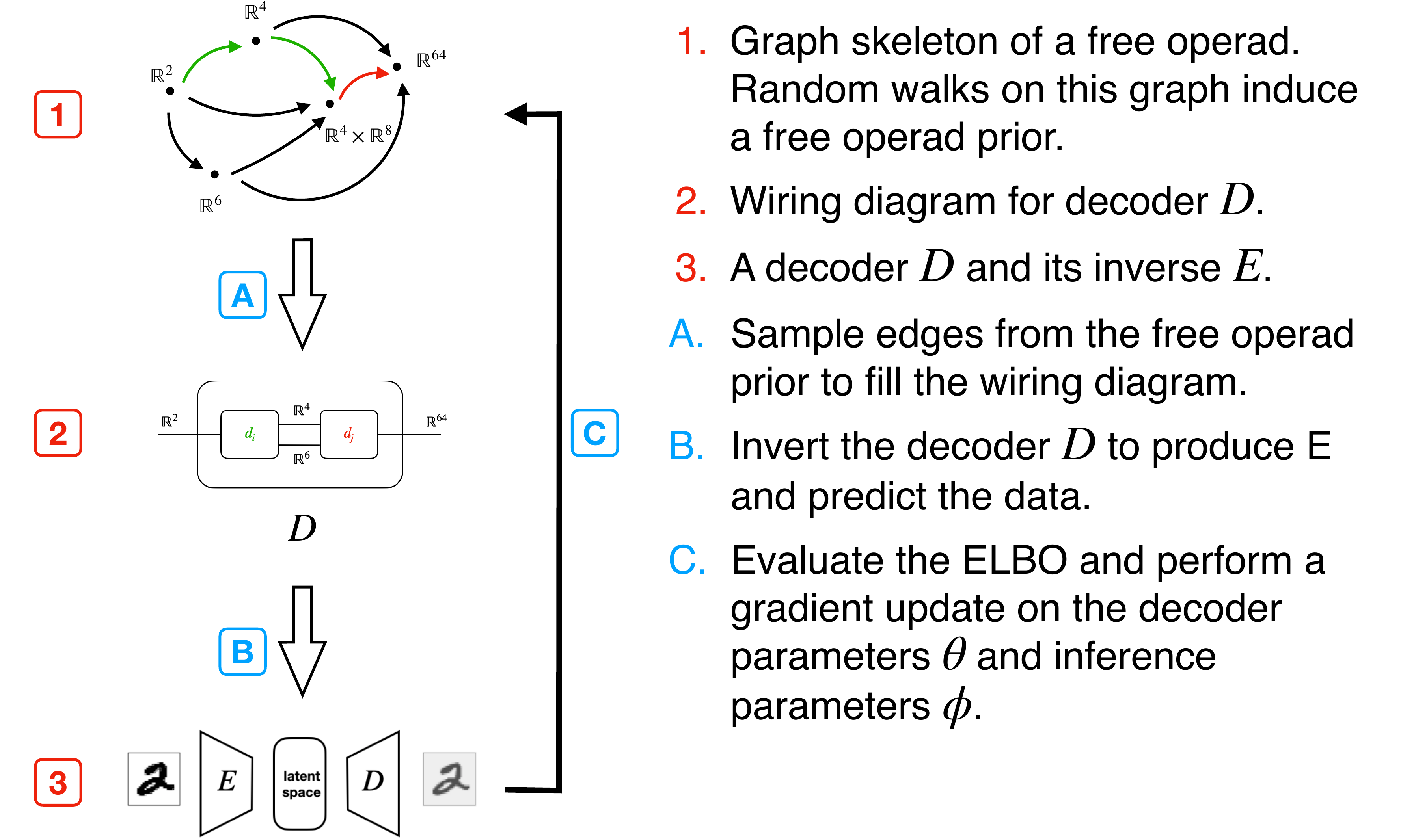}
    \caption{Example experiment. In each epoch of training, DisCoPyro learns variational autoencoder structures by sampling them from its skeleton according to a wiring diagram, then learning their faithful inverses as approximate posteriors.}
    \label{fig:training}
\end{figure*}

Category theory has been applied in various domains of mathematical science, allowing us to discover universal principles unifying diverse mathematical phenomena and thereby enabling knowledge transfer between them~\cite{Fong2019}. Applications to machine learning have been pursued recently~\cite{Shiebler2021}; however there is still a large gap between foundational mathematics and applicability in concrete machine learning tasks. This work begins filling the gap. We introduce the categorical structure learning framework DisCoPyro, a probabilistic generative model with amortized variational inference. We both provide mathematical foundations and compare with other neurosymbolic models on an example application.


Here we describe why we believe that DisCoPyro could contribute, in the long run, to developing human-level artificial general intelligence. Human intelligence supports graded statistical reasoning \cite{Lake2017}, and evolved to represent spatial (geometric) domains before we applied it to symbolic (algebraic) domains. Symmetric monoidal categories provide a mathematical framework for constructing both symbolic computations (as in this paper) and geometrical spaces (e.g.~\cite{Milnor1959}). We take Lake~\cite{Lake2017}'s suggestion to represent graded statistical reasoning via probability theory, integrating neural networks into variational inference for tractability. In terms of applications, we get competitive performance (see Subsection~\ref{experiment} below) by variational Bayes, without resorting to reinforcement learning of structure as with modular neural networks~\cite{Kirsch2018,Rosenbaum2018}.

The rest of the paper is organized as follows. In Section \ref{foundation}, we first introduce mathematical foundations of DisCoPyro (Subsections \ref{foundation} and \ref{inference}). In Section~\ref{training} we then explain how to train DisCoPyro on a task (Subsection~\ref{inference}) and provide experimental results and performance comparisons (Subsection \ref{experiment}). Figure~\ref{fig:training} demonstrates the flow of execution during the training procedure for the example task. We conclude and discuss further applications in Section~\ref{conclusions}. We provide an example implementation at \href{https://github.com/neu-pml/discopyro}{https://github.com/neu-pml/discopyro} with experiments at \href{https://github.com/esennesh/categorical\_bpl}{https://github.com/esennesh/categorical\_bpl}. DisCoPyro builds upon Pyro~\cite{bingham2019pyro} (a deep universal probabilistic programming language), DisCoCat~\cite{coecke2010mathematical} (a distributional compositional model for natural language processing~\cite{coecke2010mathematical}), and the DisCoPy~\cite{de2020discopy} library for computing with categories.



\vspace{-1em}
\subsection{Notation}
This paper takes symmetric monoidal categories (SMCs) \(\mathcal{C}\) and their corresponding operads \(\mathcal{O}\) as its mathematical setting. The reader is welcome to see Fong~\cite{Fong2019} for an introduction to these. SMCs are built from objects \(Ob(\mathcal{C})\) and sets of morphisms \(\mathcal{C}(\ty{1}, \ty{2})\) between objects $\ty{1}, \ty{2} \in Ob(\mathcal{C})$. Operads are built from types \(Ty(\mathcal{O})\) and sets of morphisms \(\mathcal{O}(\ty{1}, \ty{2})\) between types \(\ty{1}, \ty{2} \in Ty(\mathcal{O})\). An SMC is usually written \((\mathcal{C}, \otimes, I)\) with a product operation $\otimes$ over objects and morphisms and a unit $I$ of $\otimes$. In both settings, every object/type $\ty{}$ has a unique identity morphism $id_{\ty{}}$. Categories support composition $g \circ f$ on morphisms, and operads support indexed composition $g\circ_{i} f$ (for $i \in \mathbb{N}$) on morphisms.






\vspace{-1em}
\section{Foundations of DisCoPyro}\label{foundation}
\vspace{-1em}
In essence, Definition~\ref{def:signature} below exposes a finite number of building blocks (generators) from an SMC, and the morphisms constructed by composing those generators with $\circ$ and $\otimes$. For example, in categories of executable programs, a finitary signature specifies a domain-specific programming language.
\begin{definition}[Finitary signature in an SMC]
\label{def:signature}
Given a symmetric monoidal category (SMC) $\mcC$ with the objects denoted by $Ob(\mcC)$, a finitary \emph{signature} \(\mcS=(O, M)\) in that SMC consists of
\begin{itemize}
    \item A finite set $O \subseteq Ob(\mcC)$; and
    \item A finite set $M$ consisting of elements $m: \mcC(\ty{1}, \ty{2})$ for some $\ty{1}, \ty{2} \in O$, such that $\forall \ty{} \in O, m \neq id_{\ty{}}$.
\end{itemize}
\end{definition}

The following free operad over a finitary signature represents the space of all possible programs synthesized from the above building blocks (generators specified by the finitary signature). Employing an operad rather than just a category allows us to reason about composition as nesting rather than just transitive combination; employing an operad rather than just a grammar allows us to reason about both the inputs and outputs of operations rather than just their outputs.

\begin{definition}[Free operad over a signature]
\label{def:signature_operad}
The free operad $\mcO_{\mcS}$ over a signature $\mcS=(O, M)$ consists of
\begin{itemize}
    \item A set of types (representations) \(Ty(\mcO_{\mcS})=\{I\} \cup O^{\otimes}\);
    \item For every $n \in \mathbb{N}^{+}$ a set of operations (mappings) \(\mcO_{\mcS}(\ty{0}, \ldots, \ty{n-1}; \ty{n})\) consisting of all trees with finitely many branches and leaves, in which each vertex $v$ with $n-1$ children is labeled by a generator $m(v) \in M$ such that \(\dom{m(v)}\) has product length $n-1$;
    \item An identity operation \(id_{\ty{}}: \mcO_{\mcS}(\ty{}; \ty{})\) for every \(\ty{} \in Ty(\mcO_{\mcS})\); and
    \item A substitution operator \(\circ_{i}\) defined by nesting a syntax tree \(\mcO_{\mcS}(\bm{\sigma}_1, \ldots, \bm{\sigma}_m; \ty{i})\) inside another \(\mcO_{\mcS}(\ty{1}, \ldots, \ty{n-1}; \ty{n})\) when $i \in [1...n-1]$ to produce a syntax tree \(\mcO_{\mcS}(\ty{1}, \ldots, \ty{i-1}, \bm{\sigma}_{1}, \ldots, \bm{\sigma}_{m}, \ty{i+1}, \ldots \ty{n-1}; \ty{n})\).
\end{itemize}
\end{definition}

Intuitively, free operads share a lot in common with context-free grammars, and in fact Hermida~\cite{Hermida1998} proved that they share a representation as directed acyclic hypergraphs. The definition of a signature in an SMC already hints at the structure of the appropriate hypergraph, but Algorithm~\ref{alg:operad_hypergraph} will make it explicit and add edges to the hypergraph corresponding to nesting separate operations in parallel (or equivalently, to monoidal products in the original SMC).
\begin{algorithm}[t!]
\caption{Algorithm to represent a free operad as a hypergraph. The function \texttt{chunks} partitions \texttt{ty} into sublists, each an element of the set $V$.}
\label{alg:operad_hypergraph}
\KwIn{signature $\mcS=(O, M)$}
\KwOut{hypergraph $H=(V, E)$, recursion sites $R$}
\SetKwFunction{chunks}{chunks}
\SetKwFunction{map}{map}
\SetKwFunction{pop}{pop}
\SetKwFunction{push}{push}
\SetKwFunction{sublist}{sublist}
$V \gets O$\;
$E \gets$ \map{$\lambda m. (\mathrm{dom}(m), \mathrm{cod}(m))$, M}\;
$R \gets \emptyset$\;
stack $\gets \{v \in V \mid |v| > 1\}$\;
\While{$\mathrm{stack} \neq \emptyset$}{
    ty $\gets$ \pop{$\mathrm{stack}$}\;
    inhabitants $\gets$ \map{$\lambda c. \{(\mathrm{dom}(e), c) \mid e\in E, \mathrm{cod}(e)=c \}$, \chunks{$\mathrm{ty}, V$}}\;
    \ForEach{$((d_1, c_1), \ldots, (d_k, c_k)) \in \bigotimes \mathrm{inhabitants}$}{
        \If{not \sublist{$\bigotimes_{i\in[1..k]} d_i, \mathrm{ty}$}}{
            $R \gets R \cup \{\otimes[(d_1, c_1), \ldots, (d_k, c_k)]\}$
            $E \gets E \cup \{(\bigotimes_{i\in[1..k]} d_i, \bigotimes_{i\in[1...k]} c_i)\}$\;
            \If{$d \notin V$}{
                \push{$\mathrm{stack}, d$}\;
                $V \gets V \cup \{\bigotimes_{i\in[1..k]} d_i\}$\;
            }
        }
    }
}
\Return{$(V, E), R$}
\end{algorithm}
In the hypergraph produced by Algorithm~\ref{alg:operad_hypergraph}, each vertex corresponds to a non-product type and each hyperedge has a list of vertices as its domain and codomain. Each such hypergraph admits a representation as a graph as well, in which the hyperedges serve as nodes and the lists in their domains and codomains serve as edges. We will use this graph representation $G\simeq H$ to reason about morphisms as paths between their domain and codomain.

We will derive a probabilistic generative model over morphisms in the free operad from this graph representation's directed adjacency matrix $A_G$.
\begin{definition}[Transition distance in a directed graph]
\label{def:intuitive_distance}
The ``transition distance'' between two indexed vertices $v_i, v_j$ is the negative logarithm of the $i, j$ entry in the exponentiated adjacency/transition matrix
\begin{align}
    \label{eq:intuitive_distance}
    d(v_i, v_j) &= -\log \left( \left[e^{A_G} \right]_{i, j} \right),
\end{align}
where the matrix exponential is defined by the series
\begin{align*}
    e^{A_G} &= \sum_{n=1}^{\infty} \frac{(A_G)^{n}}{n!}.
\end{align*}
\end{definition}

A soft-minimization distribution over this transition distance will, in expectation and holding the indexed target vertex constant, define an probabilistic generative model over paths through the hypergraph.
\begin{definition}[Free operad prior]
\label{def:free_operad_prior}
Consider a signature $\mcS=(O, M)$ and its resulting graph representation $G=(V,E)$ and recursion sites $R$, and then condition upon a domain and codomain \(\ty{-}, \ty{+} \in Ty(\mcO_{\mcS})\) represented by vertices in the graph. The \emph{free operad prior} assigns a probability density to all finite paths \(\mathbf{e}=(e_1, e_2, \ldots, e_n)\) with \(\dom{\mathbf{e}}=\ty{-}\) and \(\cod{\mathbf{e}}=\ty{+}\) by means of an absorbing Markov chain. First the model samples a ``precision'' $\beta$ and a set of ``weights'' $\rvw$
\begin{align*}
    \beta &\sim \gamma(1, 1) &
    \rvw &\sim \mathrm{Dirichlet}\left(\Vec{\mathbf{1}}^{(|M| + |R|)}\right).
\end{align*}
Then it samples a path (from the absorbing Markov chain defined in Algorithm~\ref{alg:path}) by soft minimization (biased towards shorter paths by $\beta$) of the transition distance
\begin{align}
    \label{eq:normedpolicy}
    \pi(e \in E \mid \ty{1}, \ty{2}; \beta) &:= \frac{\exp{\left( -\frac{1}{\beta} d(\cod{e}, \ty{2}) \right)}}{
        \sum_{e' \in E : \dom{e'}=\ty{1}} \exp{\left( -\frac{1}{\beta} d(\cod{e'}, \ty{2}) \right)}
    }.
\end{align}
Equation~\ref{eq:normedpolicy} will induce a transition operator $T$ which, by Theorem 2.5.3 in Latouche and Ramaswami~\cite{Latouche1999}, will almost-surely reach its absorbing state corresponding to $\ty{2}$. This path can then be filled in according to Algorithm~\ref{alg:generator}. The precision $\beta$ increases at each recursion to terminate with shorter paths. We denote the induced joint distribution as
\begin{align}
    p(f, \rvw, \beta; \ty{-}, \ty{+}) &= p(f \mid \beta, \rvw; \ty{-}, \ty{+}) p(\rvw) p(\beta).
\end{align}
\end{definition}

\begin{algorithm}[t!]
\caption{The Markov chain constructing paths between types}\label{alg:path}
\SetKwProg{Fn}{function}{}{end}
\SetKwFunction{Generator}{Generator}
\KwData{hypergraph $(V, E)$}
\Fn{Path($\ty{-}, \ty{+}, \beta, \rvw$)}{
    $i \gets 1$\;
    $\ty{i} \gets \ty{-}$\;
    $f \gets id_{ty{-}}$\;
    \While{$\ty{i} \neq \ty{+}$}{
        $e_i \sim \pi(e \in E \mid \ty{i}, \ty{+}\; \beta)$\;
        $\ty{i} \gets \cod{e_i}$\;
        $f \gets f \fatsemi$ \Generator{$e_i, \beta, \rvw$}\;
        $i \gets i + 1$\;
    }
    \Return{f}
}
\end{algorithm}

\begin{algorithm}[t!]
\caption{Filling in an edge in the path with a morphism}
\label{alg:generator}
\SetKwProg{Fn}{function}{}{end}
\SetKwFunction{Path}{Path}
\KwData{hypergraph $(V, E)$, generators $M$, recursion sites $R$}
\Fn{Generator($e, \beta, \rvw$)}{
    $gs \gets \{m\in M \mid (\dom{m}, \cod{m}) = (\dom{e}, \cod{e}) \}$\;
    $gs \gets gs \cup \{ \otimes[(d_1, c_1), \ldots, (d_k, c_k)] \in R \mid \bigotimes_{i\in[1...k]} d_i = \dom{e} \wedge \bigotimes_{i\in[1...k]} c_i = \cod{e} \}$\;
    \ForEach{$j \in \{1, \ldots, |gs|\}$}{
        \If{$gs_j = \otimes(\ldots)$}{
            $\rvw_j \gets \rvw_j / \beta$\;
        }
    }
    $\rvw_e \gets [\rvw_{n} \mid g \in gs, g \in M, n=\mathtt{index}(g, M)]$\;
    $\rvw_e \gets \rvw_{e} + [\rvw_{|M|+n} \mid g \in gs, g \in R, n=\mathtt{index}(g, R)]$\;
    $j \sim \mathrm{Discrete}(\rvw_{e})$\;
    \eIf{$gs_j = \otimes[(d_1, c_1), \ldots, (d_k, c_k)]$}{
            \Return{$\bigotimes_{l=1}^{k}$ \Path{$d_l, c_l, \beta + 1, \rvw$}}
    }{
        \Return{$gs_j$}
    }
}
\end{algorithm}

Having a probabilistic generative model over operations in the free operad over a signature, we now need a way to specify a structure learning problem. Definition~\ref{def:wiring_diagram} provides this by specifying what paths to sample (each box specifies a call to Algorithm~\ref{alg:path}) and how to compose them. 
\begin{definition}[Wiring diagram]
\label{def:wiring_diagram}
An acyclic, $O$-typed \emph{wiring diagram}~\cite{Spivak2013,Fong2019} is a map from a series of \emph{internal boxes}, each one defined by its domain and codomain pair $(\ty{i}^{-}, \ty{i}^{+})$  to an \emph{outer box} defined by domain and codomain $(\ty{n}^{-}, \ty{n}^{+})$
\begin{align*}
    \Phi &: \mcO_{\mcS}(\ty{1}^{-}, \ty{1}^{+}) \times \ldots \times \mcO_{\mcS}(\ty{n-1}^{-}, \ty{n-1}^{+}) \rightarrow \mcO_{\mcS}(\ty{n}^{-}, \ty{n}^{+}).
\end{align*}
Acyclicity requires that connections (``wires'') can extend only from the outer box's domain to the domains of inner boxes, from the inner boxes codomains to the outer box's codomain, and between internal boxes such that no cycles are formed in the directed graph of connections between inner boxes.
\end{definition}
Given a \textbf{user-specified} wiring diagram $\Phi$, we can then wite the complete prior distribution over all latent variables in our generative model.
\begin{align}
    \label{eq:generative}
    p(f, \rvw, \beta; \Phi, \mcS) &= p(\beta) p(\rvw) \prod_{(\ty{i}^{-}, \ty{i}^{+}) \in \Phi} p(f_{i} \mid \beta, \rvw; \ty{i}^{-}, \ty{i}^{+}).
\end{align}
If a user provides a likelihood \(p_{\vtheta}(\rvx, \rvz \mid f)\) relating the learned structure $f$ to data $\rvx$ (via latents $\rvz$) we will have a joint density
\begin{align}
    \label{eq:genjoint}
    p(\rvx, \rvz, f, \rvw, \beta; \Phi, \mcS) &= p(\rvx \mid f) p(f, \rvw, \beta; \Phi, \mcS),
\end{align}
and Equation~\ref{eq:genjoint} then admits inference from the data $\rvx$ by Bayesian inversion
\begin{align}
\label{eq:bayesinverse}
    p(\rvz, f, \rvw, \beta \mid \rvx; \Phi, \mcS) &= \frac{
        p(\rvx, \rvz, f, \rvw, \beta; \Phi, \mcS)
    }{
        p_{\vtheta}(\rvx; \Phi, \mcS)
    }.
\end{align}
Section~\ref{inference} will explain how to approximate Equation~\ref{eq:bayesinverse} by stochastic gradient-based optimization, yielding a maximum-likelihood estimate of $\vtheta$ and an optimal approximation for the parametric family $\vphi$ to the true Bayesian inverse.

\vspace{-1em}
\subsection{Model learning and variational Bayesian inference}\label{inference}
Bayesian inversion relies on evaluating the model evidence $p_{\vtheta}(\rvx; \Phi, \mcS)$, which typically has no closed form solution. However, we can transform the high-dimensional integral over the joint density into an expectation
\begin{align*}
    p_{\vtheta}(\rvx; \Phi, \mcS) &= \int p_{\vtheta}(\rvx, \rvz, f_{\vtheta}, \rvw, \beta; \Phi, \mcS) d\rvz\: df_{\vtheta}\: d\rvw\: d\beta \\
    &= \mathbb{E}_{p(f_\vtheta, \rvw, \beta; \Phi, \mcS)} \left[ p_{\vtheta}(\rvx, \rvz, f_{\vtheta}, \rvw, \beta; \Phi, \mcS) \right],
\end{align*}
and then rewrite that expectation into one over the proposal
\begin{multline*}
    \mathbb{E}_{p(f_\vtheta, \rvw, \beta; \Phi, \mcS)} \left[ p_{\vtheta}(\rvx, \rvz, f_{\vtheta}, \rvw, \beta; \Phi, \mcS) \right]
    = \\
    \mathbb{E}_{q_{\vphi}(\rvz, f_{\vtheta}, \rvw, \beta \mid \rvx; \Phi, \mcS)} \left[ \frac{p_{\vtheta}(\rvx, \rvz, f_{\vtheta}, \rvw, \beta; \Phi, \mcS)}{q_{\vphi}(\rvz, f_{\vtheta}, \rvw, \beta \mid \rvx; \Phi, \mcS)} \right].
\end{multline*}
For constructing this expectation, DisCoPyro provides both the functorial inversion described in Section~\ref{example} and an amortized form of Automatic Structured Variational Inference~\cite{ambrogioni2021automatic} suitable for any universal probabilistic program.

Jensen's Inequality says that expectation of the log density ratio will lower-bound the log expected density ratio
\begin{multline*}
    \mathbb{E}_{q_{\vphi}(\rvz, f_{\vtheta}, \rvw, \beta \mid \rvx; \Phi, \mcS)} \left[ \log \frac{p_{\vtheta}(\rvx, \rvz, f_{\vtheta}, \rvw, \beta; \Phi, \mcS)}{q_{\vphi}(\rvz, f_{\vtheta}, \rvw, \beta \mid \rvx; \Phi, \mcS)} \right]
    \leq \\
    \log \mathbb{E}_{p(f_\vtheta, \rvw, \beta; \Phi, \mcS)} \left[ p_{\vtheta}(\rvx, \rvz, f_{\vtheta}, \rvw, \beta; \Phi, \mcS) \right],
\end{multline*}
so that the left-hand side provides a lower bound to the true model evidence
\begin{align*}
    \Ls(\vtheta, \vphi) = \mathbb{E}_{q_{\vphi}(\rvz, f_{\vtheta}, \rvw, \beta \mid \rvx; \Phi, \mcS)} \left[ \log \frac{p_{\vtheta}(\rvx, \rvz, f_{\vtheta}, \rvw, \beta; \Phi, \mcS)}{q_{\vphi}(\rvz, f_{\vtheta}, \rvw, \beta \mid \rvx; \Phi, \mcS)} \right]
    &\leq
    \log p_{\vtheta}(\rvx; \Phi, \mcS).
\end{align*}
Maximizing this \emph{evidence lower bound} (ELBO) by Monte Carlo estimation of its values and gradients (using Pyro's built-in gradient estimators) will estimate the model parameters $\vtheta$ by maximum likelihood and train the proposal parameters $\vphi$ to approximate the Bayesian inverse (Equation~\ref{eq:bayesinverse}) \cite{kingma2013auto}.

\vspace{-1em}
\section{Example application and training}\label{training}
\vspace{-1em}
The framework of connecting a morphism to data via a likelihood with intermediate latent random variables allows for a broad variety of applications. This section will demonstrate the resulting capabilities of the DisCoPyro framework. Section~\ref{example} will describe an example application of the framework to deep probabilistic program learning for generative modeling. Section~\ref{experiment} that describe application's performance as a generative model.

\vspace{-1em}
\subsection{Deep probabilistic program learning with DisCoPyro}\label{example}
As a demonstrative experiment, we constructed an operad $\mcO$ whose generators implemented Pyro building blocks for deep generative models $f_\vtheta$ (taken from work on structured variational autoencoders~\cite{kingma2013auto,Rezende2016,Zhao2017}) with parameters $\vtheta$. We then specified the one-box wiring diagram \(\Phi: (I, \mathbb{R}^{28\times 28}) \rightarrow (I, \mathbb{R}^{28\times 28})\) to parameterize the DisCoPyro generative model. We trained the resulting free operad model on MNIST just to check if it worked, and on the downsampled ($28 \times 28$) Omniglot dataset for few-shot learning \cite{Lake2019} as a challenge. Since the data $\rvx \in \R^{28 \times 28}$, our experimental setup induces the joint likelihood
\begin{align*}
    p_{\vtheta}(\rvx \mid \rvz, f_\vtheta) &= \mathcal{N}(\vmu_{\vtheta}(\rvz, f_{\vtheta}), \mI \vtau) \\
    p_{\vtheta}(\rvx, \rvz \mid f_{\vtheta}) &= p_{\vtheta}(\rvx \mid \rvz, f_{\vtheta}) p_{\vtheta}(\rvz \mid f_{\vtheta}).
\end{align*}
DisCoPyro provides amortized variational inference over its own random variables via neural proposals $q_{\vphi}$ for the ``confidence'' $\beta \sim q_{\vphi}(\beta \mid \rvx)$ and the ``preferences'' over generators $\rvw \sim q_{\vphi}(\rvw \mid \rvx)$. Running the core DisCoPyro generative model over structures $f_\vtheta$ then gives a proposal over morphisms in the free operad, providing a generic proposal for DisCoPyro's latent variables
\begin{align*}
    q_{\vphi}(f_{\vtheta}, \rvw, \beta \mid \rvx; \Phi, \mcS) &= p(f_{\vtheta} \mid \rvw, \beta; \Phi, \mcS) q_{\vphi}(\beta \mid \rvx) q_{\vphi}(\rvw \mid \rvx).
\end{align*}
Since the morphisms in our example application are components of deep generative models, each generating morphism can be simply ``flipped on its head'' to get a corresponding neural network design for a proposal. We specify that proposal as \(q_{\vphi}(\rvz \mid \rvx, f_{\vtheta})\); it constructs a faithful inverse~\cite{Webb2018} compositionally via a dagger functor (for further description of Bayesian inversion as a dagger functor, please see Fritz~\cite{Fritz2020}). Our application then has a complete proposal density
\begin{align}
    \label{eq:jointproposal}
    q_{\vphi}(\rvz, f_{\vtheta}, \rvw, \beta \mid \rvx; \Phi, \mcS) &= q_{\vphi}(\rvz \mid f_{\vtheta}, \rvx) q_{\vphi}(f_{\vtheta}, \rvw, \beta \mid \rvx; \Phi, \mcS).
\end{align}

\vspace{-1em}
\subsection{Experimental Results and Performance Comparison}\label{experiment}
\begin{figure*}[t]
    \centering
    \begin{subfigure}{0.45\textwidth}
        \centering
        \includegraphics[width=\textwidth]{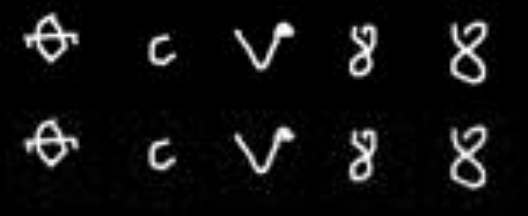}
        \caption{Omniglot characters (above) and their reconstructions (below)}
        \label{subfig:reconstructions}
    \end{subfigure}
    \hfill
    \begin{subfigure}{0.45\textwidth}
        \centering
        \includegraphics[width=\textwidth]{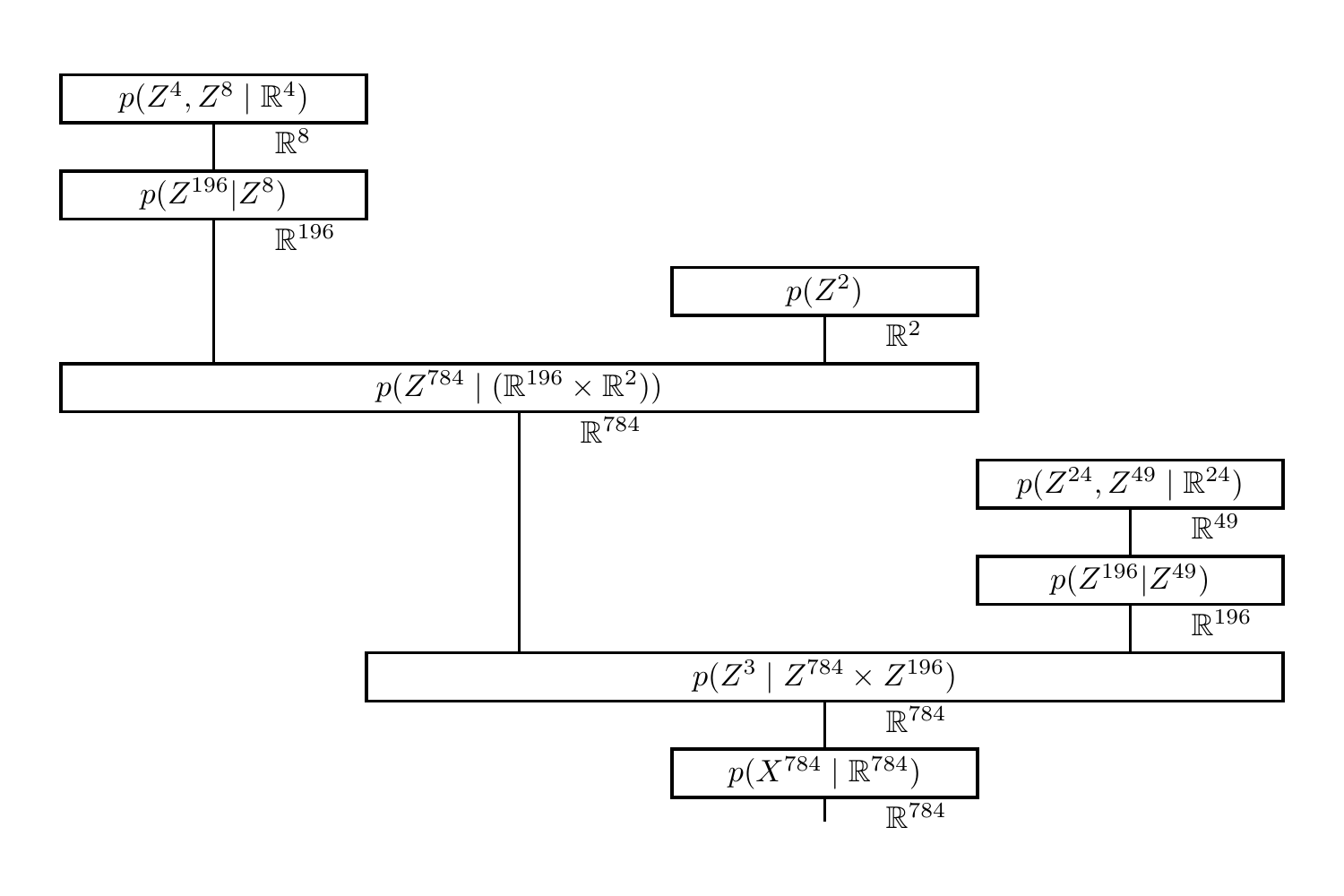}
        \caption{A string diagram sampled from the free operad model's Bayesian inverse.}
        \label{subfig:omniglotdiagram}
    \end{subfigure}
    \caption{Reconstructions (left) generated by inference in the diagrammatic generative model (right) on handwritten characters in the Omniglot evaluation set. The string diagram shows a model that generates a glimpse, decodes it into an image canvas via a variational ladder decoder, and then performs a simpler process to generate another glimpse and insert it into the canvas.}
    \label{fig:omniglotresults}
    \vspace{-1em}
\end{figure*}

\begin{table*}[t!]
    \centering
    \begin{tabular}{c|c|c|c}
        \toprule
        Model & Image Size & Learns Structure & log-$\Hat{Z}$/dim \\
        \hline 
        Sequential Attention~\cite{Rezende2016} & 28x28 & \xmark &-0.1218 \\
        Variational Homoencoder~\cite{Hewitt2018} (PixelCNN) & 28x28 & \xmark & -0.0780 \\
        Graph VAE~\cite{He2019} & 28x28 & \cmark & -0.1334 \\
        Generative Neurosymbolic~\cite{Feinman2021} & 105x105 & \cmark & -0.0348 \\
        Free Operad DGM (ours) & 28x28 & \cmark & \textbf{-0.0148} \\
        \bottomrule
    \end{tabular}
    \caption{Average log-evidence on the Omniglot evaluation set across models. Our free operad model obtains the highest higher log-evidence per data dimension.}
    \label{tab:loglikelihoods}
    \vspace{-1em}
\end{table*}
Table~\ref{tab:loglikelihoods} compares our free operad model's performance to other structured deep generative models. We report the estimated log model evidence. Our free operad prior over deep generative models achieves the best log-evidence per data dimension, although standard deviations for the baselines do not appear to be available for comparison. Some of the older baselines, such as the sequential attention model and the variational homoencoder, fix a composition structure ahead of time instead of learning it from data as we do. Figure~\ref{fig:omniglotresults} shows samples from the trained model's posterior distribution, including reconstruction of evaluation data (Figure~\ref{subfig:reconstructions}) and an example structure for that data (Figure~\ref{subfig:omniglotdiagram}).

Historically, Lake~\cite{Lake2019} proposed the Omniglot dataset to challenge the machine learning community to achieve human-like concept learning by learning a single generative model from very few examples; the Omniglot challenge requires that a model be usable for classification, latent feature recognition, concept generation from a type, and exemplar generation of a concept. The deep generative models research community has focused on producing models capable of few-shot reconstruction of unseen characters. \cite{Rezende2016} and \cite{Hewitt2018} fixed as constant the model architecture, attempting to account for the compositional structure in the data with static dimensionality. In contrast, \cite{He2019} and \cite{Feinman2021} performed joint structure learning, latent variable inference, and data reconstruction as we did.

\vspace{-1em}
\section{Discussion}\label{conclusions} 
\vspace{-1em}
This paper described the DisCoPyro system for generative Bayesian structure learning, along with its variational inference training procedures and an example application. Section~\ref{foundation} described DisCoPyro's mathematical foundations in category theory, operad theory, and variational Bayesian inference. Section~\ref{training} showed DisCoPyro to be competitive against other models on a challenge dataset.

As Lake~\cite{Lake2017} suggested, (deep) probabilistic programs can model human intelligence across more domains than handwritten characters. Beyond programs, neural network architectures, or triangulable manifolds, investigators have applied operads and SMCs to chemical reaction networks, natural language processing, and the systematicity of human intelligence \cite{phillips2010categorial,bradley2018applied}. This broad variety of applications motivates our interest in representing the problems a generally intelligent agent must solve in terms of operadic structures, and learning those structures jointly with their contents from data.

\bibliographystyle{splncs04}
\bibliography{samplepaper}
%






\end{document}